\documentclass[10pt, conference, letterpaper]{IEEEtran}
\IEEEoverridecommandlockouts
\pdfoutput=1
\usepackage{cite}
\usepackage{amsmath,amssymb,amsfonts}
\usepackage{algorithmic}
\usepackage{graphicx}
\usepackage{textcomp}
\usepackage{comment}
\usepackage{autobreak}
\usepackage{xcolor}
\usepackage{array}
\usepackage{amsmath,amssymb,amsfonts}
\usepackage{amsthm}
\usepackage{mathrsfs}
\usepackage{CJK}
\usepackage[ruled,linesnumbered]{algorithm2e}
\usepackage{amsmath}
\usepackage{url}
\usepackage{color}
\usepackage{pifont}
\usepackage{braket}
\usepackage{multirow}
\usepackage{subfigure}

\newtheorem{assumption}{Assumption}

\newtheorem{theorem}{Theorem}
\newtheorem{definition}{Definition}
\newtheorem{corollary}{Corollary}
\newtheorem{proposition}{Proposition}

\newcommand{\tabincell}[2]{\begin{tabular}{@{}#1@{}}#2\end{tabular}}

\allowdisplaybreaks[4]

\ifCLASSINFOpdf
\else
\fi
\begin{document}
%
\title{Expediting In-Network Federated Learning by Voting-Based Consensus Model Compression}

\author{\IEEEauthorblockN{Xiaoxin Su\IEEEauthorrefmark{2}, Yipeng Zhou\IEEEauthorrefmark{3}, Laizhong Cui\IEEEauthorrefmark{1} \IEEEauthorrefmark{2}\IEEEauthorrefmark{4} and Song Guo\IEEEauthorrefmark{5}}
\IEEEauthorblockA{\IEEEauthorrefmark{2}College of Computer Science and Software Engineering, Shenzhen University, Shenzhen, China}
\IEEEauthorblockA{\IEEEauthorrefmark{3}School  of Computing, Faculty of Science and Engineering, Macquarie University, Sydney, Australia}
\IEEEauthorblockA{\IEEEauthorrefmark{4}Guangdong Laboratory of Artificial Intelligence and Digital Economy (SZ), Shenzhen University, Shenzhen, China}
\IEEEauthorblockA{\IEEEauthorrefmark{5}Department of Computer Science and Engineering, The Hong Kong University of Science and Technology, HKSAR}
\IEEEauthorblockA{Email: suxiaoxin2016@163.com, yipeng.zhou@mq.edu.au, cuilz@szu.edu.cn and songguo@cse.ust.hk}

\IEEEauthorblockA{
	\thanks{
		\newline This work has been partially supported by National Key Research and Development Plan of China under Grant No. 2022YFB3102302, National Natural Science Foundation of China under Grant No. U23B2026 and No. 62372305, Shenzhen Science and Technology Program under Grant No. RCYX20200714114645048,  Key-Area Research and Development Program of Guangdong Province under Grant No. 2021B0101400003, Hong Kong RGC Research Impact Fund under Grant No. R5060-19 and No. R5034-18, Areas of Excellence Scheme under Grant AoE/E-601/22-R, General Research Fund under Grant No. 152203/20E, No. 152244/21E, No. 152169/22E and No. 152228/23E, Shenzhen Science and Technology Innovation Commission under Grant No. JCYJ20200109142008673}
		\thanks{\textit{\IEEEauthorrefmark{1}Corresponding author: Laizhong Cui}
	}
}
}

\maketitle

\begin{abstract}
Recently, federated learning (FL) has gained momentum because of its capability in preserving data privacy. To conduct model training by FL, multiple clients exchange model updates with a parameter server via Internet. To accelerate the communication speed, it has been explored to deploy a programmable switch (PS) in lieu of the parameter server to coordinate clients. The challenge to deploy the PS in FL lies in its scarce memory space, prohibiting running memory consuming aggregation algorithms on the PS. To overcome this challenge, we propose Federated Learning in-network Aggregation with Compression (FediAC) algorithm, consisting of two phases: client voting and model aggregating. In the former phase, clients report their significant model update indices to the PS to estimate global significant model updates. In the latter phase, clients upload global significant model updates to the PS for aggregation. FediAC consumes much less memory space and communication traffic than existing works because the first phase can guarantee consensus compression across clients. The PS easily aligns model update indices to swiftly complete aggregation in the second phase. Finally, we conduct extensive experiments by using public datasets to demonstrate that FediAC remarkably surpasses the state-of-the-art baselines in terms of model accuracy and communication traffic.

\end{abstract}



%
\IEEEpeerreviewmaketitle

\section{Introduction}
\label{introduction}
With the wide deployment of smart devices,  an enormous amount of data can be  collected for training advanced large models to empower intelligent services such as face recognition and natural language processing \cite{NEURIPS2021_854d9fca}, which however gives rise to privacy leakage concerns. To preserve data privacy, the federated learning (FL) paradigm has received tremendous attention which can  train models without accessing original data. FL is conducted over multiple global iterations coordinated by a parameter server. During FL training, 
clients only transmit model updates rather than original data to  the parameter server which is responsible for  aggregating clients' model updates and distributing aggregated results back to clients \cite{lim2020federated}.


Nonetheless, the communication cost is heavy if multiple clients excessively transmit model updates with the parameter server via Internet, especially when training large models with a high dimension. To diminish communication cost, it has been explored by \cite{9394352} that a programmable switch (PS) equipped with certain memory and computation capacity can take the role of the parameter server for coordinating FL clients. More importantly, the PS can achieve  a faster convergence speed \cite{10.1145/3552326.3587436} by directly  aggregating packetized model updates in the network layer.  In \cite{265065}, model updates generated by distributed machine learning are quantized  into integers which are then transmitted to the PS for aggregation. In \cite{9664035}, the PS is deployed to coordinate cross-silo FL training  across multiple institutes.  In \cite{265053}, Lao \emph{et al.} proposed  ATP attempting to leverage multiple PSes to perform decentralized, dynamic, best-effort in-network aggregation to accelerate distributed model training.

Despite the swift speed of the PS, its scare memory space is a major obstacle restricting its efficiency by straightly applying existing FL algorithms, \emph{e.g.}, FedAvg \cite{mcmahan2017communication},    in in-network FL for model aggregation.  For example, a typical PS can only allocate 1 MB  memory space for FL \cite{pan2022enabling}. By executing FedAvg, the PS can only process $2.5\times 10^5$ model updates per aggregation assuming that each model update takes 4 bytes.  It implies that it takes an excessive number of aggregations to process large models. For instance, it at least takes $4,000$ aggregations to process a large model with 1 billion parameters, which can extremely deteriorate the training efficiency of FL \cite{10.1145/3605153}. 

Model compression is  a prospective approach to shrinking communication traffic. Unfortunately, most existing works are designed for FL coordinated by parameter servers, not taking the scarce memory limitation for aggregation into account. There are mainly two types of compression algorithms: quantization and sparsification. In essence, quantization expresses each model update with a fewer number of bits such as TernGrad \cite{wen2017terngrad} and SignSGD \cite{bernstein2018signsgd}.
For in-network FL, quantization is indispensable since   the PS can only process integer numbers. In \cite{265065}, model updates are always quantized into integers before they are aggregated by the PS. 
Sparsification  will discard insignificant model updates of a small magnitude such as Topk \cite{stich2018sparsified}, DGC \cite{lin2017deep} and DC2 \cite{abdelmoniem2021dc2}. Existing sparsification techniques are more effective in shrinking communication traffic, yet a high compression rate does not necessarily result in less memory consumption on the PS. 



To fully exert the advantage of the PS for coordinating FL clients, we design the Federated Learning in-network Aggregation with Compression (FediAC) algorithm by considering the limited memory space of the PS. FediAC consists of two phases: client voting and model aggregating. The purpose of Phase 1 is to estimate global significant model updates, \emph{i.e.}, model updates of a large magnitude which can dominate the model change per global iteration. More specifically, each client uploads an array of 0s and 1s to report indices of local significant model updates to the PS. Then, the PS estimates global significant model updates by scanning uploaded arrays once to  deduce  a global index array (GIA) indicating consensus significant model updates across different clients.  The overhead cost of Phase 1 is small because each model update is only represented by a single bit. In Phase 2,  clients upload quantized model updates according to the GIA so that the PS can easily align model updates  to swiftly complete aggregation  with limited memory space. 
Moreover,  we theoretically prove  the convergence of FediAC, based on which we can properly tune the compression rate of FediAC. 
To verify the superb performance of FediAC, we conduct extensive experiments by using  CIFAR-10, CIFAR-100 and FEMNIST datasets. Experimental results demonstrate that FediAC can outperform the state-of-the-art in-network aggregation baselines by improving model accuracy 1.15\%-7.71\% or shrinking communication traffic by 41.14\%-69.91\%.

The rest of the paper is organized as follows. Related works and Preliminaries are introduced in Sec.~\ref{RelatedWork} and Sec.~\ref{Preliminaries}. We elaborate the training process of FediAC and analyze its convergence in Sec.~\ref{Analysis}. At last, we conclude our work in Sec.~\ref{Conclusion} after thoroughly conducting  experiments in Sec.~\ref{Experiment}.

\section{Related Work}
\label{RelatedWork}
In this section, we explore related works from three perspectives: FL design, model compression in FL and in-network aggregation for FL.

\noindent{\bf Federated Learning.}  FL is a distributed machine learning framework originally designed by Google to protect data privacy \cite{mcmahan2017communication} for mobile users when training the next word prediction model. 
To coordinate model training in FL, a parameter server  iteratively communicates with clients by exchanging models (or model updates) instead of raw data with decentralized clients. FedAvg is the most classical algorithm for FL model training, which adopts multi-round local training to improve communication efficiency. 
The convergence of FedAvg is theoretically guaranteed by  \cite{li2019convergence, yang2021achieving} under the non-IID data distribution. 

{\color{black}To overcome the communication bottleneck  of a single parameter server in  FL  \cite{pokhrel2020federated}, prior studies have proposed parallel and hierarchical architectures so that multiple servers can collaboratively coordinate FL clients. P-FedAvg proposed in \cite{zhong2021} completely covers all clients by deploying multiple  servers to avoid the single server failure risk in FL. Wang \emph{et al.} \cite{9488756} proposed a  hierarchical architecture which can organize multiple edge nodes into a tree to  speed up the training of FL.}

All these designs did not consider the constraint of memory space when aggregating models because they simply assume that model aggregation will be  executed by a powerful server.  {\color{black}
Moreover, the deployment cost for multiple servers is much heavy, which can be considerably reduced by deploying switches for model aggregation. 
}

\noindent{\bf Model Compression.} Model compression can be broadly divided into quantization and sparsification, which is very effective in shrinking communication traffic in FL.  Quantization  reduces the number of bits to represent each model update at the cost of lowered precision. FedHQ \cite{9425020} is a heterogeneous quantization compression algorithm in FL that assigns heterogeneous quantization errors to clients with different aggregation weights so as to improve the convergence speed of FL. Xu \emph{et al.} \cite{9288933} proposed a federated training ternary quantization (FTTQ) algorithm that utilizes a self-learning quantization factor to compress model updates in order to lower communication cost. {\color{black}However, these algorithms quantizing model updates into floating numbers are not feasible for in-network FL because PSes can only perform integer arithmetic.}  Sparsification, on the other hand, reduces the number of transmitted model updates to optimize communications in FL. {\color{black} Topk  \cite{stich2018sparsified} is the most representative sparsification algorithm, which can achieve a much higher compression rate than quantization by only transmitting top $k$ parameters. Therefore, a Topk based model compression algorithm should be developed for  in-network FL as well to achieve the highest learning speed, which is still an open problem. 
}

\noindent{\bf In-Network FL.}
With the rapid improvement of computation capacity such as GPUs \cite{jouppi2017datacenter}, it has made the network in distributed learning a key factor determining the training time cost. 
Thanks to the advent of programmable switches (PSes) \cite{10.1145/2486001.2486011}, the network burden can be alleviated through in-network aggregation. 
In particular, according to \cite{9664035},  it is feasible to deploy a PS to aggregate model updates  for cross-silo FL consisting of a number of powerful and stable clients. 
In \cite{altamore2022accelerating}, Vera Altamore incorporated programmable P4 switches into FL to compute intermediate aggregations and used edge nodes for in-network model caching and gradient aggregating to speed up the training process.
SwitchML \cite{265065} is a framework that combines in-network aggregation with distributed learning, designing the  protocol for end-hosts and the framework for machine learning to speed up training. SwitchML breaks down  model updates into  sized blocks so that the PS can process them in a pipelined manner. In addition, the framework utilizes the scoreboard mechanism of the switch and the retransmission mechanism of  end-hosts to handle packet losses.
Fang \emph{et al.} \cite{10050420} proposed the first distributed in-network aggregated gradient routing (GRID) algorithm, considerably improving the model aggregation speed by deliberately  routing model update packets to the PS with aggregation capability.

Intuitively speaking, model compression can be combined with in-network aggregation to boost  FL performance. However, all existing compression designs have not considered the constraint of limited memory space when aggregating models. A high compression rate does not necessarily imply much less memory consumption. To optimize in-network FL with compression, we contribute to developing a novel in-network FL algorithm with minimized PS  memory  consumption.


\section{Preliminaries}
\label{Preliminaries}
This section includes the introduction of  preliminary knowledge on traditional FL and a motivation example of our study.
\subsection{Traditional FL}
We briefly introduce how the FL process is coordinated by a parameter server.  
Suppose that there are $N$ clients and one parameter server, where the private dataset in client $i$ is $\mathcal{D}_i$ with candidates $D_i$. Note that the data distribution on clients is not identically and independently distributed (non-IID). The objective of FL is to train a model   $\mathbf{w}\in\mathbb{R}^d$ of dimension $d$ by minimizing a loss function, 
{\em i.e.,} $\mathbf{w}^*=\arg\min_\mathbf{w}F(\mathbf{w})=\frac{1}{N}\sum_{i=1}^NF_i(\mathbf{w}, \mathcal{D}_i).$
Here, $F_i(\mathbf{w}, \mathcal{D}_i)$ is the local loss function of client $i$ defined by $F_i(\mathbf{w}, \mathcal{D}_i)=\frac{1}{D_i}\sum_{\xi\in\mathcal{D}_i}f(\xi, \mathbf{w})$, where $\xi$ is a particular data sample and $f()$ is a function to evaluate model  $\mathbf{w}$ using $\xi$.  
Typically, FL minimizes $F(\mathbf{w})$ by iteratively conducting  $T$ global iterations. In the $t-$th global iteration, the training process is presented as follows:
\begin{itemize}
	\item The parameter server 
 distributes  the latest global model $\mathbf{w}_t$ to all clients.
	\item  Client $i$ initializes its local model as $\mathbf{w}^i_{t,0}=\mathbf{w}_t$ and performs $E$ rounds of local iterations using the batch gradient descent algorithm: $$\mathbf{w}^i_{t,j+1}=\mathbf{w}^i_{t,j}-\eta_{t,j}\nabla F_i(\mathbf{w}^i_{t,j}, \mathcal{B}^i_{t,j}),$$ where $j=0,1,\dots,E-1$ and $\mathcal{B}^i_{t,j}$ is a batch data of size $B$ sampled from $\mathcal{D}_i$. After that, client $i$ uploads the model updates $\mathbf{U}^i_t=\mathbf{w}^i_{t,0}-\mathbf{w}^i_{t,E}$ to the remote parameter server for aggregation.
	\item The server aggregates  uploaded model updates to refine the global model by: $\mathbf{w}_{t+1}=\mathbf{w}_t-\frac{1}{N}\sum_{i=1}^N\mathbf{U}^i_t$ and kick off the next global iteration.
\end{itemize}

\subsection{A Motivation Example}

Based on the FL process,  we can find that the parameter server is mainly responsible for aggregating model updates. According to \cite{10050420}, the simple aggregation operation can be completed by a programmable switch (PS) in a much faster speed if all model updates are quantized as integers. This architecture can been applied to accelerate cross-silo FL \cite{9664035}. 
Despite that the PS can accelerate FL in the network layer, the memory space of a PS is very limited, \emph{e.g.,}  $\sim 64$ MB for P4-Programmable Ethernet Switch with Intel Tofino2 chip \cite{agrawal2020intel}. 
Therefore, in-network FL is generally performed through integer rounding of model updates first, and then  pipelined aggregating of rounded model updates \cite{265065, 10.1145/3452296.3472904}.

We use a concrete example to illustrate how the limited PS memory restricts the aggregation speed. 
Suppose that there are two clients collaboratively training a model with 5 parameters. The local model updates on two clients are $\mathbf{U}_1 = [5,4,3,2,1]$ and $\mathbf{U}_2 = [1,3,4,5,2]$, respectively, after local training and quantization. If the memory space of the PS is limited to only aggregate a pair of integers per aggregation, it takes 5 aggregations for the PS to aggregate $\mathbf{U}_1$ and $\mathbf{U}_2$.

However, directly compressing models cannot effectively reduce the number of aggregations on the PS. For example, if the Top2 algorithm is applied, each client only uploads 2 top model updates to the PS implying that Client 1 uploads model updates with indices 1 and 2, while Client 2 uploads model updates with indices 3 and 4. The PS cannot align the indices to aggregate these model updates directly, which in the end takes 4 aggregations to process  4 top model updates. 

\begin{figure}[htbp]
\vspace{-0.45cm}
	\setlength{\abovecaptionskip}{-0.1cm}
	\centering
	\includegraphics[width=\linewidth]{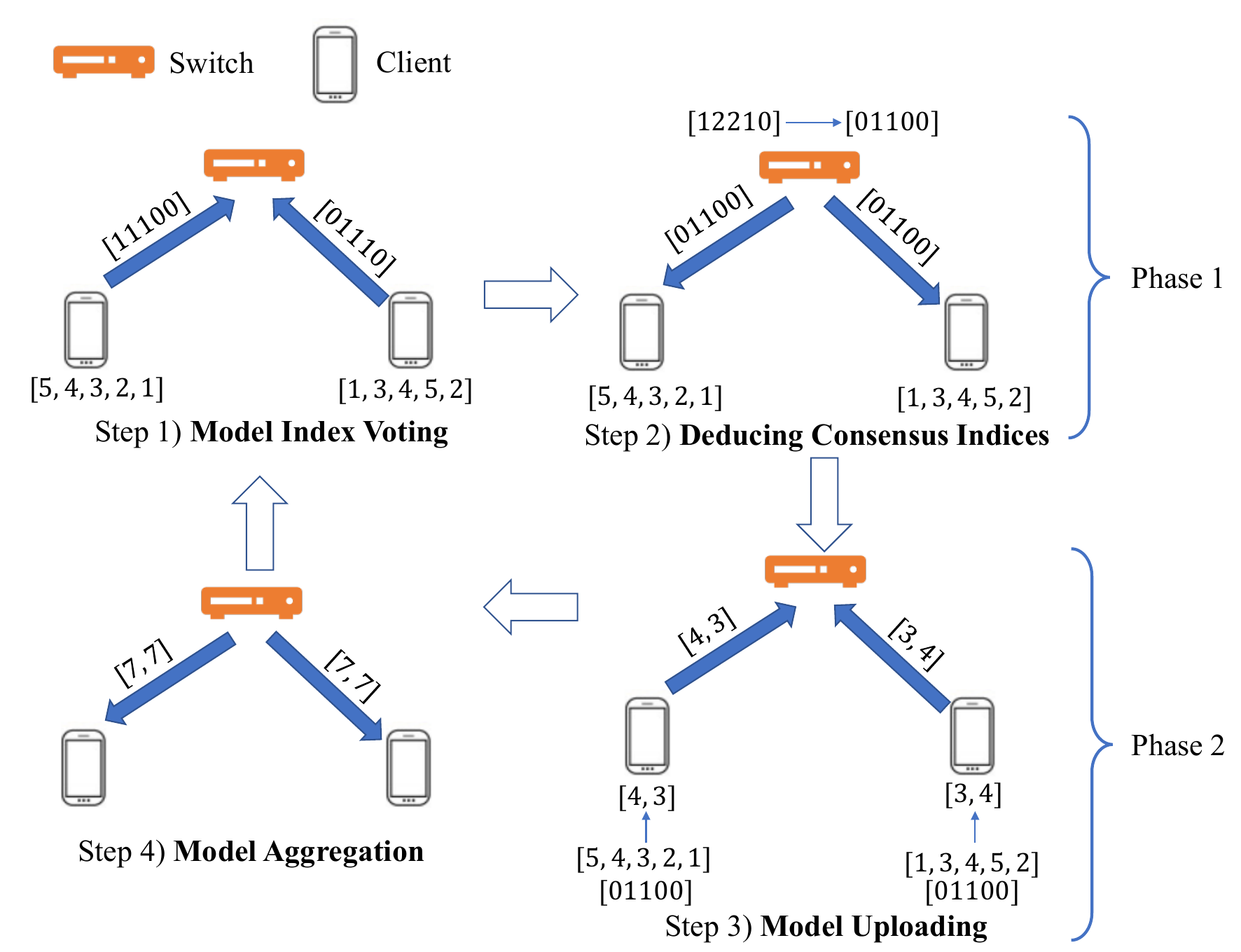}
	\caption{The training process of FediAC.}
	\label{fig:FediAC}
 \vspace{-0.2cm}
\end{figure}

Through this simple example, we can see that how fast  the indices of model updates can be aligned is crucial for the PS efficiency. Inspired by this example, we design the in-network FL with two phases, and indices of uploaded model updates will be aligned in Phase 1. Let us stick to the same example. In Phase 1, each client uploads a 0-1 array to the PS yielding $11100$ and $01110$ arrays from two clients, respectively, to indicate their top 3 model updates. The PS aggregates two arrays to get $12210$ and only reserves top 2 model updates by returning $01100$ to clients. In Phase 2, clients  upload model updates according to $01100$, \emph{i.e.}, uploading model updates with indices 2 and 3. 
Note that it only costs one aggregation by Phase 1 because it  only needs  5 bits to represent a local model in Phase 1. Therefore,  it takes 3 aggregation operations in total on the PS. The detailed process is presented in Fig.~\ref{fig:FediAC}.

This concrete example just illustrates the dilemma confronted by in-network FL. If a powerful server is adopted for model aggregation, it is easy to allocate 16 GB or more memory to conduct aggregation \cite{265065} which can easily overcome the difficulty to align model update indices for aggregation. 

\section{Algorithm Design and Analysis}
\label{Analysis}

In this section, we elaborate the design of our FediAC algorithm, and theoretically prove its convergence. At last, its implementation in practical systems is discussed. 


\subsection{FediAC Algorithm Design}
To simplify our discussion, we suppose that all clients index their model parameters and model updates in the same order from $1$ to $d$. 
Corresponding to four steps in Fig.~\ref{fig:FediAC},  FediAC  conducts the $t$-th global iteration as below:
\begin{enumerate}
	\item \textbf{Model Index Voting.} Client $i$ synchronizes its local model $\mathbf{w}^i_{t,0}$ with the global model $\mathbf{w}_{t}$ based on  the last round global model  $\mathbf{w}_{t-1}$ and aggregated model updates distributed  from the PS. 
 Then, client $i$ performs $E$ local iterations to obtain the local model $\mathbf{w}^i_{t,E}$. Next, the client gets local model updates by  $\mathbf{U}^i_t=\mathbf{w}^i_{t,0}-\mathbf{w}^i_{t,E}+\mathbf{e}^i_{t-1}$, where $\mathbf{e}^i_{t-1}$ denotes the residual error representing  the accumulated  model updates not uploaded to the PS until  global iteration $t-1$. Client $i$ probabilistically votes $k$ elements.  The odds to vote each model update is proportional to its magnitude, which will be further specified later. 
     Client $i$ just uploads a 0-1 index array $\vec{v}^i_t$ to the PS to report what model updates have been voted. 
     In the index array, ``1"/``0" represents that a particular dimension of model updates is selected/unselected. 
	\item \textbf{Deducing Consensus  Indices.} The PS aggregates  received 0-1 arrays to get $\vec{v}_t=\sum_i \vec{v}^i_t$. 
  Each element in $\vec{v}_t$ is denoted by $v_l^t$ for $l=1,\dots, d$.
 After aggregating all index arrays, the PS uses a threshold integer $a$ to filter out insignificant model updates in $\vec{v}_t$. In other words, if $v_l^t < a$, FediAC sets $v_l^t=0$. Otherwise, if $v_l^t \ge a$, FediAC sets $v_l^t=1$. 
 The implication is that a particular dimension is significant only if at least $a$ clients have voted it.  Then, the PS distributes $\vec{v}_t$, regarded as the GIA (global index array),  back to clients. 
	\item \textbf{Model Uploading.}  Each client selects  elements in $\mathbf{U}$ to upload according to $\vec{v}_t$. Since the PS  only performs integer arithmetic \cite{265065}, we quantize selected model updates in an unbiased manner before uploading. In general, we can assume that each model update element, denoted by  $U_l$,  is quantized into a $b$-bit integer as follows.  First, $U_l$ is multiplied by a factor $f$ for amplification to obtain $fU_l$, where $f=\frac{2^{b-1}-N}{Nm}$ and $m$ is the maximum absolute value of model updates \cite{265065}. The element is then quantized by:
	\begin{equation}
		\label{EQ:QuantizationEq}
		\theta(fU_l)=\left\{
		\begin{aligned}
			& \lfloor fU_l \rfloor,\ \textit{with prob.}\ \lceil fU_l \rceil - fU_l,\\
			&\lceil fU_l \rceil,\ \textit{with prob.}\ fU_l-\lfloor fU_l \rfloor. \\
		\end{aligned}
		\right.
	\end{equation}
	It is easy to verify that $\mathbb{E}[\theta(fU_l)]=fU_l$, implying that the quantization is unbiased. 
{\color{black} Then, the model update is sparsified by $\pi(\theta(fU_l))=\theta(fU_l)*v_l$ based on   $\vec{v}$ received from the PS. To make our presentation concise, let $\Theta$ (corresponding to $\theta$) and $\Pi$ (corresponding to $\pi$) denote the quantization and sparsification operations on mode update vectors. 
 Compressing model updates $\mathbf{U}^i_t$ can be expressed by $\Pi(\Theta(f\mathbf{U}^i_t))$. 
 Meanwhile,  client $i$ updates the local residual  error by $\mathbf{e}^i_t=\frac{1}{f}(f\mathbf{U}^i_t-\Pi(\Theta(f\mathbf{U}^i_t)))$ and uploads non-zero elements in $\Pi(\Theta(f\mathbf{U}^i_t))$ to the PS.}
	\item \textbf{Model Aggregation.} Finally, the PS  aggregates integer model updates by pipelined processing model updates to get $\sum_{i=1}^N\Pi(\Theta(f\mathbf{U}^i_t))$, which are sent back to all clients. Clients can update the global model by $\mathbf{w}_{t+1}=\mathbf{w}_t-\frac{1}{Nf}\sum_{i=1}^N\Pi(\Theta(f\mathbf{U}^i_t))$ prior to  the next global iteration.
\end{enumerate}

The pseudocode  of FediAC is  presented  in Algo.~\ref{algo:FediACAlgorithm}.
\begin{algorithm}[t]
	\label{algo:FediACAlgorithm}
	\caption{Training process of FediAC}
	\LinesNumbered
	 \KwIn{residual error $\mathbf{e}^i_0=\mathbf{0}$, initial global model $\mathbf{w}_1$.}
	 \KwOut{global model $\mathbf{w}_{T+1}$.}
    \For{$t=1,2,\dots,T$}{
	\textbf{On client $\boldsymbol{i}$:}\\
	Initialize local model $\mathbf{w}^i_{t,0}=\mathbf{w}_t$ and update model $\mathbf{w}^i_{t,j}=\mathbf{w}^i_{t,j-1}-\eta_t\nabla F_i(\mathbf{w}^i_{t,j-1},\mathcal{B}^i_{t,j-1})$ for $E$ iterations to get $\mathbf{w}^i_{t,E}$.\\
	Compute model updates $\mathbf{U}^i_t=\mathbf{w}^i_{t,0}-\mathbf{w}^i_{t,E}+\mathbf{e}^i_{t-1}$.\\
    \tcp{Model Index Voting Step.}
	Vote $k$ elements in  $\mathbf{U}^i_t$ using odds proportional to their magnitude to get $\Vec{v}^i_t$. \\
	Upload voting array $\Vec{v}^i_t$ to PS.\\
    Download GIA $\vec{v}_t$ from PS.\\
    \tcp{Model Uploading Step.}
	Use $\Vec{v}_t$ to select model updates and perform unbiased integer quantization using \eqref{EQ:QuantizationEq}.\\
    Update residual error $\mathbf{e}^i_t=\frac{1}{f}(f\mathbf{U}^i_t-\Pi(\Theta(f\mathbf{U}^i_t)))$.\\
	Upload non-zero elements in $\Pi(\Theta(f\mathbf{U}^i_t))$ to PS.\\
	Download aggregated data $\sum_{i'=1}^N\Pi(\Theta(f\mathbf{U}^{i'}_{t}))$ from PS.\\
    Update global model $\mathbf{w}_{t+1}=\mathbf{w}_{t}-\frac{1}{Nf}\sum_{i'=1}^N\Pi(\Theta(f\mathbf{U}^{i'}_{t}))$.\\
	\textbf{On PS:}\\
    \tcp{Deducing Consensus Indices Step.}
	Aggregate 0-1 arrays from clients to get $\vec{v}_t=\sum_i\vec{v}^i_t$ and deduce a consensus indices according to the threshold $a$.\\
	Distribute GIA to clients.\\
    \tcp{Model Aggregation Step.}
	Aggregate uploaded data from clients to get $\sum_i\Pi(\Theta(f\mathbf{U}^i_t))$.\\ 
    Distribute aggregated model updates to clients.\\
 }
\end{algorithm}
Note that uploaded model updates will be encapsulated into multiple packets for Internet communications from clients to the PS. Since the indices of model updates have been aligned by $\vec{v}_t$, it is also easy to align the indices of model updates encapsulated in the same packet as long as each client encapsulates the same number of model updates into a single packet. The PS can swiftly aggregate model updates by  adding communication packets in a pipelined manner. 

In FediAC, there are two critical parameters influencing the aggregation cost of the PS, which are the voting threshold $a$ and the number of bits $b$ for quantization.  In FediAC, $a$ is regarded as a hyperparameter  to adjust the compression rate.  If $a $ is larger, more model updates will be filtered out implying a higher compression rate. The value of  $b$ determines the quantization error and the cost  to represent each model update. If the quantization error is too large, it can make FL diverge. Thus, there is a lower bound of $b$ to guarantee the convergence of FL, which will be derived by convergence analysis. 



\subsection{Compression Error Analysis}

The cost of compression operations in FediAC lies in lowered model update precision. If the compression error is giant, it is possible that FL training diverges in the end. To guarantee the convergence of FediAC, we analyze the compression error bound of  FediAC to guarantee convergence.

FediAC compresses model updates with both quantization and sparsification operations.  Sparsification  removes insignificant model updates and the significance of a model update is formally defined by its magnitude. 
The work \cite{m2021efficient} reported that model update magnitudes obey a power-law distribution. Similarly, to simplify our analysis, we assume that the magnitudes of model updates on different clients can be uniformly bounded by a power-law distribution \cite{stich2018sparsified}. 
\begin{definition}
	\label{DEF:PowerDist}
	For client $i$, model updates ranked in a descending order of their absolute values satisfy a power law distribution, \emph{i.e.,} $|U^i_t\{l\}|\le \phi l^{\alpha}, \quad \textit{for } l=1,\dots,d$ and $t =1,\dots, T$. Here, we let  $U^i_t\{l\}$ denote the $l$-th largest model update (in terms of absolute value) in $\mathbf{U}^i_t$. $\alpha<0$ is the decay exponent controlling the decaying rate of the distribution and $\phi$ is a constant. 
\end{definition}
 Based on the power-law distribution, the magnitude of the $l$-th largest model update is $\phi l^{\alpha}$.
All clients vote their local model updates according to the power-law distribution. To ease our discussion, let $\vec{v}_t\{l\}$ denote the index of the model update corresponding to $U^i_t\{l\}$. 
According to the FediAC design, the global model update process can be expressed by 
$\mathbf{w}_{t+1}=\mathbf{w}_t-\frac{1}{Nf}\sum_{i=1}^N\Pi(\Theta(f\mathbf{U}^i_t)),$
where $\mathbf{U}^i_t$ represents model updates obtained from local training, $\Theta()$ is an unbiased integer quantization operation and $\Pi()$ represents the  sparsification function conducted by clients based on the received GIA $\vec{v}_t$ from the PS. Here, $f$ is the scaling factor to mitigate quantization error.

In step 1) {\bf Model Index Voting}, client $i$ votes $k$ elements with a probability proportional to its model update magnitude. Based on Definition~\ref{DEF:PowerDist}, client $i$ 
voting $k$ model updates will select the $l$-th largest model update with probability:
\begin{align}
    \label{EQ:SamplePro}
    p_{l}=\frac{l^{\alpha}}{\sum_{l'=1}^d(l')^{\alpha}}.
\end{align}
Given \eqref{EQ:SamplePro}, the probability to vote $U^i_t\{l\}$ by client $i$  is:
\begin{align}
    \label{EQ:VotedPro}
    q_{l}=1-(1-p_{l})^k.
\end{align}

In step 2) {\bf Deducing Consensus Indices}, the PS filters out insignificant model updates if there are fewer than $a$ clients voting for them.  Therefore, we can calculate the probability that $\Vec{v}_t\{l\}=1$ in the GIA vector $\vec{v}_t$ returned by PS as:
\begin{align}
    \label{EQ:UploadedPro}
    r_l=\sum_{j=a}^{N}\tbinom{N}{j}(q_l)^j(1-q_l)^{N-j}.
\end{align}
Then, the expected number of model updates that can be uploaded by FediAC clients is $\mathbb{E}[k_S]=\sum_{l=1}^{d}r_l.$

For analyzing the compression error in  $\Pi(\Theta(f\mathbf{U}^i_t))$, the complexity mainly lies in the sparsification operation, \emph{i.e.}, $\Pi$. 
Based on \eqref{EQ:UploadedPro}, we derive the compression error of FediAC as follows. 

\begin{proposition}
	\label{PRO:ErrorBound}
	In FediAC, each client votes a model update in $\mathbf{U}^i_t$ with the odds proportional to its magnitude. When a model update is voted by at least $a$ clients, the 
 model update is scaled up by $f$ and  quantized to an integer of $b$ bits by \eqref{EQ:QuantizationEq}. Then, the compression error of FediAC is bounded by $\mathbb{E}\|\Pi(\Theta(f\mathbf{U}^i_t))-f\mathbf{U}^i_t\|^2\le\gamma\|f\mathbf{U}^i_t\|^2$. Here,
	\begin{align}
		\label{EQ:ErrorBound}
		\gamma=1-\frac{\sum_{l=1}^dr_ll^{2\alpha}}{\sum_{l=1}^dl^{2\alpha}}+\frac{1}{4f^2}\frac{\sum_{l=1}^dr_l}{\phi^2\sum_{l=1}^dl^{2\alpha}},
	\end{align}
	where $f=\frac{2^{b-1}-N}{Nm}$ and $m$ is the maximum  value of  model update magnitudes. Recall that $(f\mathbf{U}^i_t)\{l\}$ is the $l$-th largest element 
 in $f\mathbf{U}^i_t$ and $r_l$ is the probability to upload $(f\mathbf{U}^i_t)\{l\}$ by clients.
\end{proposition}

Intuitively speaking, the compression error is derived as the sum of the error caused by discarding insignificant model updates in $\Pi()$ and the error caused by quantization in \eqref{EQ:QuantizationEq}. 
The detailed proof is presented in Appendix~\ref{Proof:ProErrorBound}. 

\noindent{\bf Remark.} It is worth noting that both $a$ and $b$ influence the compression error according to \eqref{EQ:ErrorBound}. If $a$ is too large, an excessive number of model updates will be filtered out. Likewise, if $b$   is too small, it implies that    each model update is only expressed by a very small number of  bits resulting in a large quantization error. To make FediAC converges, we should avoid setting  $a$ over-large or $b$ over-small.

\begin{corollary}
If $a$ is fixed, to ensure the convergence of FediAC, it is necessary to set $b$ by 
\begin{align}
	\label{EQ:ConstrainOfb}
	b>\log_2\left(\frac{\sqrt{\sum_{l=1}^dr_l}}{2\phi\sqrt{\sum_{l=1}^dr_ll^{2\alpha}}}*Nm+N\right) + 1.
\end{align}
\end{corollary}
The lower bound of $b$ is derived by letting $0<\gamma<1$, which is a necessary condition to make FL converge \cite{9589061}. It will be further validated by subsequent convergence analysis. 


\subsection{Convergence Analysis}

We move on to prove that FediAC can converge under non-convex loss as long as $0< \gamma <1$. 
In order to analyze the convergence of FediAC, we make the following conventional assumptions frequently used in previous works \cite{li2019convergence, dinh2020federated}.
\begin{assumption}(L-Smoothness)
	\label{Assump:ConSmoo}
	All local loss functions, \emph{i.e.}, $F_1, F_2,\dots, F_N$ are  $L$-smooth. In other words, given $\mathbf{w}_1$ and $\mathbf{w}_2$, we have $F_i(\mathbf{w}_1) \le F_i(\mathbf{w}_2) + (\mathbf{w}_1-\mathbf{w}_2)^T\nabla F_i(\mathbf{w}_2)+\frac{L}{2}\|\mathbf{w}_1-\mathbf{w}_2\|^2$, for $i=1,\dots,N$.
\end{assumption}
Let $\xi^i_{t,j}$ denote any sample randomly sampled from $\mathcal{D}_i$ in the $j$-th local iterations of the $t$-th global iteration.
\begin{assumption}
	\label{Assump:LocalVar}
	The variance of the stochastic gradients in each client is bounded, \emph{i.e.},
	$\mathbb{E}[\|\nabla F_i(\mathbf{w}^i_{t,j}, \xi^i_{t,j})-\nabla F_i(\mathbf{w}^i_{t,j})\|^2] \le \sigma^2$ for $\forall i,j,t$.
\end{assumption} 
\begin{assumption}
	\label{Assump:BoundG}
	The expected square norm of stochastic gradients is uniformly bounded, \emph{i.e.},  $\mathbb{E}[\|\nabla F_i(\mathbf{w}^i_{t,j}, \xi^i_{t,j})\|^2] \le G^2$ for $\forall i,j,t$.
\end{assumption}

It is likely that the data distribution among FL clients is not independently and identically distributed (non-IID). According to \cite{wang2019adaptive}, the degree of non-IID   can be defined as follows.
\begin{definition} (Quantification of non-IID)
	\label{QuantificationOfnon-IID}
	The difference between global gradients and local gradients, \emph{i.e}, $\mathbb{E}[\|\nabla F_i(\mathbf{w}_t)-\nabla F(\mathbf{w}_t)\|^2] \le {\color{black}\Gamma}^2, \forall i,t$, is used to quantify the degree of non-IID of the sample distribution among clients.
\end{definition}



FediAC performs model training for $T$ global iterations, where the learning rate at the $t-$th global iteration is $\eta_t$. We define $H=\sum_{t=1}^{T}\eta_t$ and all clients participate in FL. Let $\mathbf{x}$ denote a global model sampled based on the learning rate weights, \emph{i.e.,} $Pr(\mathbf{x}=\mathbf{w}_t)=\frac{\eta_t}{H}$. 
Based on  Proposition~\ref{PRO:ErrorBound} and  assumptions, we derive the following theorem.
\begin{theorem}
	\label{Theo:ConverRate}
	Let Assumptions~\ref{Assump:ConSmoo}-\ref{Assump:BoundG} hold. The learning rate is set as $\eta_t=\frac{1}{\sqrt{t+q}}$ (constant $q>0$) which satisfies $\frac{1}{2}-15E^2\eta_t^2L^2\ge0$. The compression error satisfies $0<\gamma<1$.   Let $a$, $b$ and $f$ denote the voting threshold, the number of bits for quantization and the scaling factor, respectively. After conducting $T$ global iterations with $N$ clients, the convergence rate of FediAC is bounded by:
	\begin{align}
		\|\nabla F(\mathbf{x})\|^2&\!\!\le\!\!\frac{F(\mathbf{w}_1)-F^*+\delta}{cE\sqrt{T}} +\frac{L^2EG^2\frac{\gamma}{(1-\sqrt{\gamma})^2}}{2c\sqrt{T}},\notag
	\end{align}
	where $F^*$ is the global loss computed by optimal model parameters $\mathbf{w}^*$, $c =\frac{1}{2}-15E^2\eta_t^2L^2\ge0$, $\delta=\frac{L+1}{2}E^2G^2+\frac{5E^2L^2}{2}(\sigma^2+6E{\color{black}\Gamma}^2)$ and  $\gamma$ is defined by Proposition~\ref{PRO:ErrorBound}.
\end{theorem}
It can be proved based on the theoretical derivation  in \cite{9589061}, in which the convergence of FL is derived with a biased compression operation. In \cite{9589061}, it is required that the compression error must satisfy $0 < \gamma < 1$, which can be guaranteed by properly setting $a$ and $b$ in Proposition~\ref{PRO:ErrorBound}. Thereby, the convergence proof can be completed by substituting $\gamma$ derived in  Proposition~\ref{PRO:ErrorBound} into the convergence rate in \cite{9589061}. Due to limited space, the detailed proof is omitted to avoid repetition.

\subsection{Implementation of FediAC}
We discuss two issues on implementation of the FediAC algorithm in practical in-network FL systems. 

\noindent{\bf How to tune $\boldsymbol{a}$ and $\boldsymbol{b}$.}
Although we have theoretically analyzed how $a$ and $b$ influence the compression  error and the convergence rate of FediAC, it is not easy to tune $a$ and $b$ centrally by the PS. We discuss how to set them in a lightweight manner in practice. 
A parameter server with more powerful capacity than the PS can easily  assist in setting $a$ and $b$ by participating in the first  global iteration. In the first  global iteration,  clients report all model updates to the server since $\vec{v}_1$ is unknown. Once receiving all these model updates, the server can fit the power-low distribution in reported model updates to obtain $\alpha$ and $ \phi$, based on which the server can further set $a$ and $b$. For $a$, it can be tuned like a hyperparameter by simulating different values of $a$ to search the best one on the server. For each given value of  $a$,  $b$ is set according to \eqref{EQ:ConstrainOfb} to minimize the load on the PS.

In global iteration $t=2$, the server  broadcasts aggregated model updates, $a$ and $b$ to clients and the PS, and  will not participate in FL anymore. Clients  keep using $a$ and $b$ fixed by the server in the rest iterations over the entire training process. 

\noindent{\bf Overhead of Phase 1.}
Compared with existing works, Phase 1 in  FediAC brings the unique advantage, which however  incurs overhead.  This overhead has been minimized by FediAC using a single bit to represent a model dimension. For a large model with 10 million model updates, it only takes 1.25 MB traffic. 
For extremely high-dimension models such as ChatGPT with billions of parameters, we should explore compression techniques such as  run-length encoding compression techniques (which are particularly effective in compressing 0-1 arrays) \cite{ZHANG2021526} to  shrink the size of  index arrays in Phase 1.

\section{Performance Evaluation}
\label{Experiment}
In this section, we simulate a PS to process model aggregation for multiple FL clients. Based on our simulator, we carry out extensive experiments to validate the superb performance of FediAC in comparison with the state-of-the-art baselines. 

\subsection{Experimental Settings}
\subsubsection{Datasets and Models}
\begin{itemize}
	\item Both {\bf CIFAR-10} and {\bf CIFAR-100} are datasets consisting of 60,000 color images which can be classified into 10 and 100 classes, respectively. In each dataset, 50,000 images are used as the training set and the remaining 10,000 images are used as the test set.
    Each image in both datasets is 3*32*32 in size. We allocate images to clients obeying  IID and non-IID distributions, respectively. For IID,  samples of the entire training set are shuffled and uniformly divided among all clients. Therefore, the label distribution is the same for different clients. For non-IID, we use the Dirichlet distribution 
    to generate label distributions on different clients and assign data samples to each client based on the generated label distributions. Note that the default parameter of the Dirichlet distribution denoted by $\beta$  is set to 0.5 \cite{li2021model}. 
	\item {\bf FEMNIST} is one of the benchmark datasets for evaluating FL performance, which consists  of 28*28 grayscale images. The dataset has 62 labels consisting of uppercase letters, lowercase letters and numbers. FEMINIST contains  handwritten images generated by different users, and its data distribution  is inherent non-IID. The number of images distributed on an individual client is 300-400.
\end{itemize}

We use the classical ResNet-18 model \cite{he2016deep}, containing  10+ million parameters, to classify the CIFAR-10 and CIFAR-100 datasets, respectively. For the FEMNIST dataset, we train a 2-layer CNN for classification, where each layer consists of Convolution-BatchNormalization-MaxPooling. The CNN is finally terminated by 3 fully connected layers. The model contains about 800,000 parameters. Based on the convergence analysis of FediAC, we set the learning rate as $\frac{0.1}{1+\sqrt{t}/40}$ and $\frac{0.1}{1+\sqrt{t}/20}$ for training ResNet-18 and CNN, respectively, where $t$ is the number of global iterations.

\subsubsection{System Settings}
By default, we set up a single PS to coordinate 20 clients for conducting FL, to simulate a cross-silo FL scenario according to \cite{kairouz2021advances}. In each global iteration, all clients  participate in the training, and each client conducts $E=5$ local iterations to update the model. After local training, model updates are encapsulated into packets which are then transmitted to the PS. In our experiment,  the default size of each packet is 1,500 bytes \cite{islam2016quality}. 

We implement queuing models to simulate the process for the PS to aggregate model updates. 
More specifically, we implement  
M/G/1 models in the FL system. 
Each client uploads model updates following a Poisson process with the rate determined by its network transmission rate. 
The arrival of packets on the PS also follows  a Poisson process. The service time to process an aggregation by the PS follows a general distribution. Let $\lambda_i$ denote the packet delivery rate of client $i$. Then, the Poisson process at the PS is parameterized by $\lambda_s=\sum_{i=1}^N\lambda_i$.
We also define the expected value and variance of the processing time for a packet aggregated by the PS as $\rho$ and $\varrho$, respectively. 
Similarly, we also implement  M/G/1 queuing models to simulate the process of each client for downloading and updating global models. The download process of each client follows the Poisson process with rate $\lambda_c$. The processing time for updating can follow a general distribution. 
To overcome the randomness of queuing models, we suppose that the PS in all algorithms has sufficient space to temporarily cache packets before they are aggregated. 


We use the traces recorded in \cite{8638955} for the cellular network in New York City to set the uploading rates of FL clients for uploading model update packets. These traces were collected from scenarios of subway traveling in New York City, recording packets sent at each timestamp in milliseconds. We use these traces to calculate the rate at which packets are sent per second and assign the calculated rate to each client. Uploading rates per second generated by the traces for different clients  range from 200 to 2,800 packets. Considering that the download speed is much higher than the upload speed, we define that the download speed of the PS is 5 times of the average uploading rates of clients in our experiments. 

The PS  conducts the aggregation operation with received model update packets.   We implement two different switches: the high performance PS and the low performance PS. The time cost per aggregation is $3.03*10^{-6}$s and $3.03*10^{-7}$s for low and high performance PSes, respectively, to process each packet. 
The variance of the time for each aggregation 
is $2.15*10^{-8}$. 
In addition, we adopt a Gaussian distribution to model the processing delay of each packet in our experiment. For the time consumed in training the model locally on clients, we set it to 0.1s for FEMNIST, 2s for CIFAR-10, and 3s for CIFAR-100, respectively.

\subsubsection{Baselines}
We implement the state-of-the-art in-network FL aggregation algorithms for comparison including libra, OmniReduce and SwitchML. Both libra and OmniReduce are designed for sparse networks, and thus  model updates obtained from training will be sparsified using Topk before uploading. Each baseline is briefly introduced as follows:
\begin{itemize}
	\item {\bf libra \cite{pan2022enabling}} divides model parameters into hot and cold types, representing parameters that will be updated frequently and rarely, respectively. 
	The switch is only responsible for the aggregation of hot parameters. Cold parameters are redirected to a remote server for aggregation.
	\item {\bf OmniReduce \cite{10.1145/3452296.3472904}} divides the model updates into different packets and only uploads the packets with non-zero elements to the PS for aggregation.
	\item {\bf SwitchML \cite{265065}} can reduce the amount of traffic transmitted by adjusting the number of bits per parameter after  quantizing model parameters into integers.
\end{itemize}

We tune hyperparameters in baselines and select best values in our experiments for comparison. For libra and OmniReduce, which are based on Topk compression, we enumerate $k$ as $0.1\%d$, $0.5\%d$, $1\%d$ and $5\%d$. 
The experimental results show that the optimal $k$ for libra and OmniReduce is $1\%d$ and $5\%d$, respectively. {\color{black}Note that libra pretrains models by a parameter server 
to predict the types of parameters. This overhead is not included in our experiments.} 
For SwitchML,  
we enumerate the number of bits for quantization as 8, 10, 12 and 14, and finally set the best  12 bits for quantization. Therefore, we compare FediAC with the optimal performance achieved by baselines.

In  FediAC, each client votes $k$ local model updates, but the total number of finally selected clients is mainly determined by the voting threshold $a$ implying that $k$ is not a critical parameter in FediAC. Thus,  we simply set $k=5\%d$ in our experiments. 
For  threshold $a$, {\color{black}we enumerate 1, 2, 3 and 4. According to our experimental results,  we set $a$ to 3 for the FEMNIST, CIFAR10\_IID and CIFAR100\_IID scenarios, and 4 for the CIFAR10\_non-IID and CIFAR100\_non-IID scenarios. Given $a$, each client determines the quantization bit $b$ based on local model updates. Specifically, each client calculates $\alpha$ and $\phi$ of the power-law distribution by using local model updates at the first global iteration. Then  $a$ is substituted into \eqref{EQ:ConstrainOfb} to set $b$.
}

\subsection{Experimental Results}

We  compare FediAC with other algorithms from two perspectives: model accuracy on test sets and consumed communication traffic to reach target accuracy.

\subsubsection{Comparing Model Accuracy}
We compare model accuracy on test sets obtained by different algorithms after conducting FL over  the same  training time duration. The experiment results are presented in Fig.~\ref{fig:ModelAccu}, in which the x-axis represents elapsed training time and the y-axis represents model accuracy on test sets. From the experimental results, we can get the following observations:

\begin{figure*}[htbp]
	\setlength{\abovecaptionskip}{-0.2cm}
	\centering
	\includegraphics[width=0.95\linewidth]{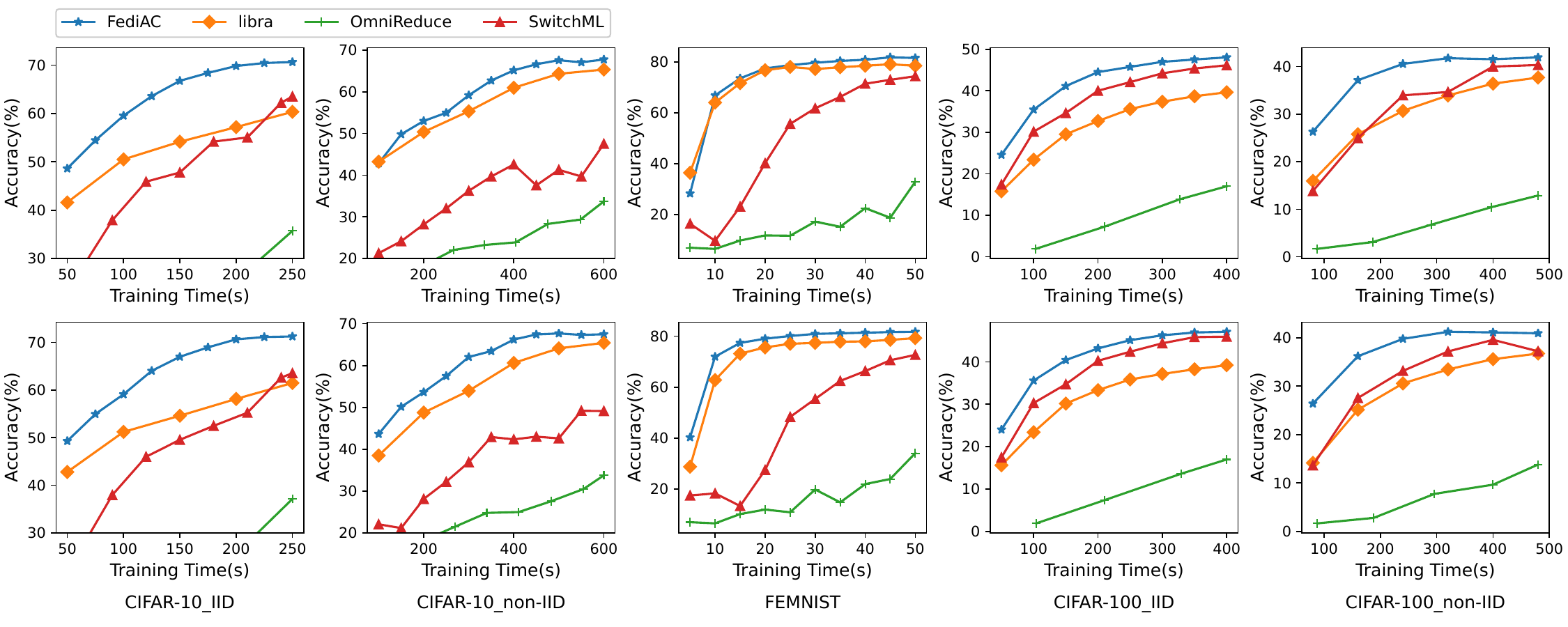}
	\caption{Comparing model accuracy with difference datasets and data distributions with a high performance (top) PS and a low performance (bottom) PS.}
	\label{fig:ModelAccu}
 \vspace{-0.6cm}
\end{figure*}

\begin{itemize}
	\item 
Compared with baselines, FediAC is always the best one to achieve the fastest convergence speed in terms of the wall-clock time regardless of the PS performance. FediAC can reach the highest final model accuracy, and its performance consistently outperforms baselines by 
improving model  accuracy from 1.15\% to 7.71\% on three datasets under both IID and non-IID data distributions.
	\item There is no  algorithm that can reliably achieve the second best performance.  SwitchML is the second best one  on  CIFAR-100, whereas libra is the second best one on CIFAR-10 and FEMNIST.
	\item OmniReduce is the worst one because of its design, which will upload   a packet as long as a single non-zero element exists  in the packet. Therefore the compression rate of OmniReduce is limited,  failing to    attain high model accuracy in a limited training time span.
    \item The performance gain achieved by  FediAC on FEMNIST is not very significant because the model trained for classifying FEMNIST images is relatively simple with only 800,000 model parameters.

\end{itemize}




\subsubsection{Comparing Communication Traffic}
We further compare the total communication traffic of the entire FL system, including both upload and download traffic over the entire training process, between FediAC and the second best baseline algorithm. The execution of an algorithm lasts until   the model reaches target  accuracy on test set. We set the target model accuracy according to datasets and data distributions. Specifically, the target accuracy is 63\% for CIFRA-10\_IID, 64\% for CIFAR-10\_non-IID, 79\% for FEMNIST, 45\% for CIFAR-100\_IID and 41\% for CIFAR-100\_non-IID. 

The experimental results are shown in Tables~\ref{tab:CommTrawithHigh} for the high performance PS and \ref{tab:CommTrawithLow} for the low performance PS, respectively. Note  that we only compared FediAC with second best baseline algorithm because  other baselines cannot reach the target accuracy  at all. Based on the experimental results, we  can observe that FediAC always consumes the least amount of total communication traffic, and can shrink 41.14\%-69.91\% of communication traffic compared to the second best baseline, manifesting that FediAC can considerably alleviate the network traffic  burden in FL.


\begin{table}[h]
\vspace{-1.0em}
\setlength{\abovecaptionskip}{-0.1em}
	\centering
	\caption{Comparing total communication traffic (upload+download traffic) consumed by different algorithms to reach target model accuracy with the high performance PS.}
	\begin{tabular}{|c|c|c|c|}
		\hline
		& \multicolumn{2}{c|}{Algo. \& Traffic} & Reduced \% \\
		\hline
		CIFAR-10\_IID(63\%) & \tabincell{c}{FediAC\\1,732 MB}   & \tabincell{c}{SwitchML\\5,756 MB}  & 69.91\%\\
		\hline
		CIFAR-10\_non-IID(64\%) & \tabincell{c}{FediAC\\4,169 MB}   & \tabincell{c}{libra\\12,219 MB}  & 65.88\%\\
		\hline
		FEMNIST(79\%) & \tabincell{c}{FediAC\\528 MB}   & \tabincell{c}{libra\\897 MB}  & 41.14\%\\
		\hline
        CIFAR-100\_IID(45\%) & \tabincell{c}{FediAC\\3,658 MB}   & \tabincell{c}{SwitchML\\7,707 MB}  & 52.54\%\\
		\hline
		CIFAR-100\_non-IID(41\%) & \tabincell{c}{FediAC\\3,550 MB}   & \tabincell{c}{SwitchML\\10,276 MB}  & 65.45\%\\
		\hline
	\end{tabular}
	\label{tab:CommTrawithHigh}
 \vspace{-1em}
\end{table}

\begin{table}[h]
\setlength{\abovecaptionskip}{-0.1em}
	\centering
	\caption{Comparing total communication traffic (upload+download traffic) consumed by different algorithms to reach target model accuracy with the low performance PS}
	\begin{tabular}{|c|c|c|c|}
		\hline
		& \multicolumn{2}{c|}{Algo. \& Traffic} & Reduced \% \\
		\hline
		CIFAR-10\_IID(63\%) & \tabincell{c}{FediAC\\1,763 MB}   & \tabincell{c}{SwitchML\\5,756 MB}  & 69.37\%\\
		\hline
		CIFAR-10\_non-IID(64\%) & \tabincell{c}{FediAC\\4,037 MB}   & \tabincell{c}{libra\\12,127 MB}  & 66.72\%\\
		\hline
		FEMNIST(79\%) & \tabincell{c}{FediAC\\352 MB}   & \tabincell{c}{libra\\995 MB}  & 64.62\%\\
		\hline
        CIFAR-100\_IID(45\%) & \tabincell{c}{FediAC\\3,539 MB}   & \tabincell{c}{SwitchML\\7,707 MB}  & 54.08\%\\
		\hline
		CIFAR-100\_non-IID(41\%) & \tabincell{c}{FediAC\\3,547 MB}   & \tabincell{c}{SwitchML\\11,561 MB}  & 69.32\%\\
		\hline
	\end{tabular}
	\label{tab:CommTrawithLow}
 \vspace{-2.0em}
\end{table}


\subsubsection{Robustness Evaluation}
\begin{figure}[htbp]
	\setlength{\abovecaptionskip}{-0.1cm}
    \centering
    \includegraphics[width=0.95\linewidth]{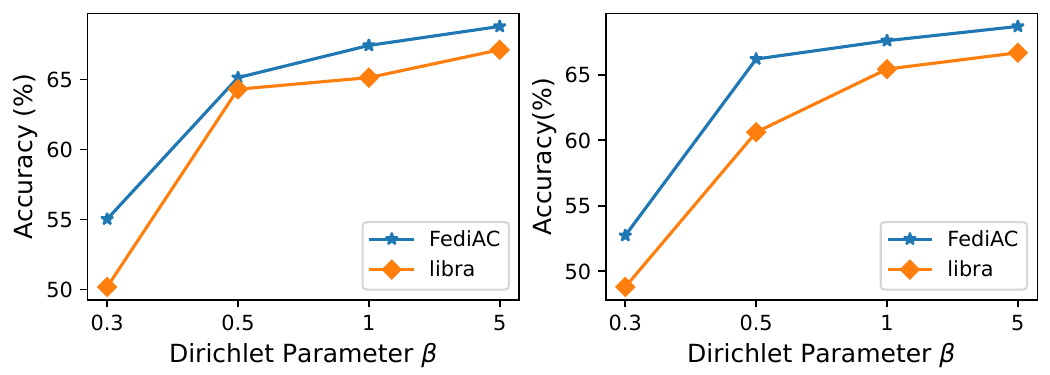}
    \caption{Comparing model accuracy of FediAC and libra by varying the  non-IID degree in CIFAR-10  with the high performance PS (left) and the low performance PS (right).}
    \label{fig:nonIIDDegree}
    \vspace{-0.4cm}
\end{figure}

To evaluate the robustness of FediAC, we vary the non-IID data degree and the system scale in our experiments. Due to limited space, we just conduct robustness evaluation by using the CIFAR-10 dataset.

When varying the non-IID data degree, we adjust the parameter $\beta$ of the Dirichlet distribution from $0.3$ to $5$. A smaller $\beta$ implies a  stronger  non-IID degree, and vice verse. {\color{black}Each algorithm is set up with a training time of 500s.}  Since  libra is the second best one in the CIFAR-10\_non-IID scenario, we only compare FediAC with libra in this experiment. The experimental results are shown in Fig.~\ref{fig:nonIIDDegree}, in which the x-axis represents the Dilichlet distribution parameter $\beta$ and the y-axis represents the final model accuracy. 

From Fig.~\ref{fig:nonIIDDegree}, we can see that the final model accuracy gradually improves as the non-IID degree decreases. Regardless of the non-IID degree, FediAC  always outperforms  libra  when using either the high or low performance PS. 

\begin{figure}
	\setlength{\abovecaptionskip}{-0.1cm}
    \centering
    \includegraphics[width=0.95\linewidth]{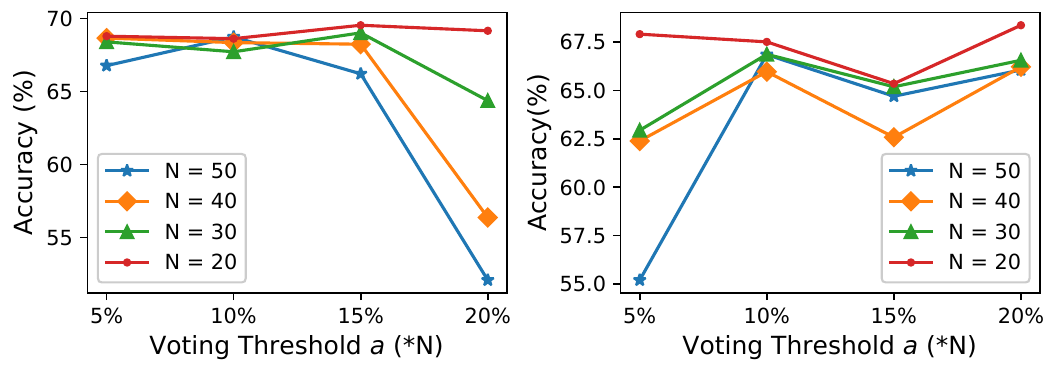}
    \caption{Comparing model accuracy trained by FediAC using different hyperparameter $a$ under different system scales with IID (left) and non-IID (right) data distributions in CIFAR-10.}
    \label{fig:systemScale}
    \vspace{-0.5cm}
\end{figure}

We further evaluate the performance of FediAC under different system scales by varying the number of clients $N$. 
Meanwhile, we enumerate the voting threshold $a$ from $5\%N$ to $20\%N$. We still use the CIFAR-10 dataset plus the low performance PS for this experiment. The settings of data distributions of CIFAR-10 in this experiment are the same as those in Fig.~\ref{fig:ModelAccu}. For different experiment cases, we simply fix  the same training time, {\em i.e.,} 500s.
The experimental results are presented in Fig.~\ref{fig:systemScale}, in which the x-axis represents different voting thresholds $a$ (denoted by a percentage of the client population $N$) and the y-axis represents the final model accuracy. We vary $N$ from $20$ to $50$. 
According to the experimental results, we can observe that:
\begin{itemize}
    \item $a$ is a critical parameter that can substantially influence the final model accuracy. If $a$ is chosen from a proper range, the performance of FediAC is very stable no matter how we change the system scale. 
  Specifically, if $a\in [5\%N, 15\%N]$  under the IID data distribution or $a\in [10\%N, 20\%N]$ under the non-IID data distribution, FediAC approximately achieves the highest model accuracy.
This property is appealing implying that it is not difficult to tune $a$ in practice. 
  \item Since we keep the total training time span fixed for experiments with different $N$, the model accuracy degrades as we increase $N$ because it takes a longer time to complete a round of global iteration if $N$ is larger. 
   

\item Under IID scenarios, it is better to set a smaller $a$ because it is easier to reach the consensus on significant model updates. In contrast, non-IID scenarios need more client votes to reach the consensus on the significance of a model update.  
\end{itemize}

\section{Conclusion}
\label{Conclusion}
The proliferation of large models aggravates the communication challenge of FL. To speed up the FL process, in-network aggregation executed by the PS emerges which can considerably improve the model aggregation speed for FL. However, a PS can only perform  integer arithmetic  with very limited memory space, raising the challenge on redesign of model aggregation algorithms in accordance with the characteristics of a PS.  To overcome the limited memory space challenge, we propose a novel FediAC algorithm to compress model updates by both quantization, converting all model updates into integer numbers, and sparsification, discarding  insignificant model updates. In particular, FediAC adopts a consensus sparsification operation across different clients so that model aggregation can be swiftly completed by the PS  to surpass existing baselines. With the advances of computation capacity, it is envisioned that large models will be more prevalent in the future, calling for solutions to alleviate  communication burden. Our initial research on in-network FL only considers a single PS in the system. Considering the potential large-scale of networks, we will extend our algorithm to make it applicable for   FL systems with multiple collaborative PSes.

\appendices
\section{Proof of Proposition~\ref{PRO:ErrorBound}}
\label{Proof:ProErrorBound}
	In FediAC, the probability that the element with the $l-$th largest absolute value in  model updates $\mathbf{U}^i_t$ is selected for uploading is $r_l$. Selected elements are quantified as  integers by \eqref{EQ:QuantizationEq} in an unbiased manner. Thus, the compression error is:
	\begin{align}
		&\mathbb{E}\frac{\|\Pi(\Theta(f\mathbf{U}^i_t))-f\mathbf{U}^i_t\|^2}{\|f\mathbf{U}^i_t\|^2}\notag\\
		&=\mathbb{E}\frac{\sum_{l=1}^d[r_l[\theta(fU^i_t\{l\})-fU^i_t\{l\}]^2+(1-r_l)(fU^i_t\{l\})^2]}{\sum_{l=1}^d(fU^i_t\{l\})^2}\notag\\
		&=\mathbb{E}\frac{\sum_{l=1}^dr_l[(\theta(fU^i_t\{l\})-fU^i_t\{l\})^2-fU^i_t\{l\}^2]}{\sum_{l=1}^d(fU^i_t\{l\})^2}+1\notag\\
		&\overset{(a)}{\le}\mathbb{E}\frac{\sum_{l=1}^dr_l[0.25-(fU^i_t\{l\})^2]}{\sum_{l=1}^d(fU^i_t\{l\})^2}+1\notag\\
		&\overset{(b)}{\approx} 1-\frac{\sum_{l=1}^dr_ll^{2\alpha}}{\sum_{l=1}^dl^{2\alpha}}+\frac{1}{4f^2}\frac{\sum_{l=1}^dr_l}{\phi^2\sum_{l=1}^dl^{2\alpha}}=\gamma.
	\end{align}
	Inequality $(a)$ is obtained based on the following analytical derivation in \eqref{EQ:BoundTheta} and $(b)$ is because we use the power-law distribution to analyze the model updates in Definition~\ref{DEF:PowerDist}. Given a floating number $U$ and its quantized result $\theta(U)$ according to the quantization formula in \eqref{EQ:QuantizationEq}, we get:
	\begin{align}
 \label{EQ:BoundTheta}
		&\mathbb{E}[\theta(U)-U]^2-U^2\notag\\
		&=(\lceil U\rceil-U)^2(U-\lfloor U\rfloor)+(\lfloor U\rfloor-U)^2(\lceil U\rceil-U)-U^2\notag\\
		&=(U-\lfloor U\rfloor)(\lceil U\rceil-U)-U^2\le 0.25-U^2.
	\end{align}

\clearpage

\bibliographystyle{IEEEtran} 
\bibliography{reference}
	
\end{document}